\documentclass{article}

\usepackage{microtype}
\usepackage{graphicx}
\usepackage{wrapfig}
\usepackage{subcaption}
\usepackage{booktabs} 
\usepackage[T1]{fontenc}

\usepackage{hyperref}

\usepackage[accepted]{icml2024}

\usepackage{amsmath}
\usepackage{amssymb}
\usepackage{mathtools}
\usepackage{amsthm}
\usepackage[normalem]{ulem}

\usepackage[capitalize,noabbrev]{cleveref}

\theoremstyle{plain}

\theoremstyle{definition}

\theoremstyle{remark}

\usepackage[textsize=tiny]{todonotes}

\icmltitlerunning{Deconstructing the Goldilocks Zone}

\definecolor{burntorange}{rgb}{0.8, 0.33, 0.0}

\begin{document}

\twocolumn[
\icmltitle{Deconstructing the Goldilocks Zone of Neural Network Initialization}

\begin{icmlauthorlist}
\icmlauthor{Artem Vysogorets}{cds}
\icmlauthor{Anna Dawid}{fi}
\icmlauthor{Julia Kempe}{cds,courant}
\end{icmlauthorlist}

\icmlaffiliation{cds}{Center for Data Science, New York University, 60 Fifth Ave, New York, NY 10011}
\icmlaffiliation{fi}{Center for Computational Quantum Physics, Flatiron Institute, 162 Fifth Ave, New York, NY 10010, USA}
\icmlaffiliation{courant}{Courant Institute, New York University, 251 Mercer St, New York, NY 10012}

\icmlcorrespondingauthor{Artem Vysogorets}{amv458@nyu.edu}

\icmlkeywords{Goldilocks Zone, Initialization, Curvature}

\vskip 0.3in
]

\printAffiliationsAndNotice{} 

\begin{abstract}
The second-order properties of the training loss have a massive impact on the optimization dynamics of deep learning models. \citet{goldilocks} discovered that a large excess of positive curvature and local convexity of the loss Hessian is associated with highly trainable initial points located in a region coined the ``Goldilocks zone''. Only a handful of subsequent studies touched upon this relationship, so it remains largely unexplained. In this paper, we present a rigorous and comprehensive analysis of the Goldilocks zone for homogeneous neural networks. In particular, we derive the fundamental condition resulting in excess of positive curvature of the loss, explaining and refining its conventionally accepted connection to the initialization norm. Further, we relate the excess of positive curvature to model confidence, low initial loss, and a previously unknown type of vanishing cross-entropy loss gradient. To understand the importance of excessive positive curvature for trainability of deep networks, we optimize fully-connected and convolutional architectures outside the Goldilocks zone and analyze the emergent behaviors. We find that strong model performance is not perfectly aligned with the Goldilocks zone, calling for further research into this relationship.
\end{abstract}

\section{Introduction}
\label{Sec:Introduction}
Every neural network gives rise to a high-dimensional optimization space spanned by its trainable parameters. The complex geometry of these spaces and embedded loss landscapes has been an area of prolific research since the inception of machine learning models. The training loss Hessian and its properties have received a lot of attention, as they offer key insights into generalization \citep{flat, keskar}, convergence speed \citep{convergence1}, and broader optimization dynamics \citep{break, eos}.

\begin{figure}[h]
\centering
\includegraphics[width=\linewidth]{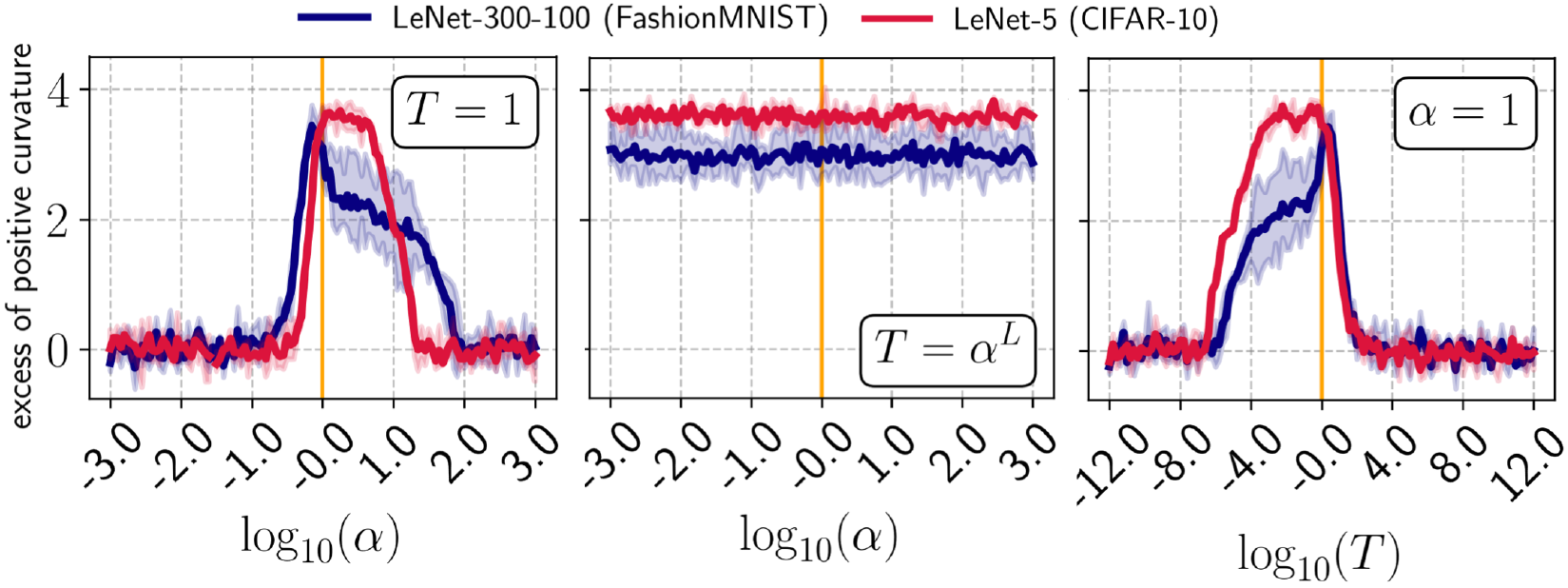}
\caption{The Goldilocks zone is an area of excess of positive curvature of the loss. \textbf{Left:} Originally, the Goldilocks zone is observed for a narrow range of initialization scales. \textbf{Middle:} Setting the appropriate softmax temperature $T$ allows for excess positive curvature at initialization of any norm. \textbf{Right:} Recreating the Goldilocks zone at an unscaled initialization just by varying $T$.}
\label{Fig:IntroGoldilocks}
\end{figure}

While the Hessian is extensively studied throughout training and at convergence, fewer works focus on the initialization stage. Recently, \citet{goldilocks} discovered the \emph{Goldilocks zone}---a region of the optimization space marked by excessive positive curvature $\text{Tr}(H)/\lVert H\rVert_F$ and local convexity of the loss Hessian $H$ (see Figure \ref{Fig:IntroGoldilocks}). The anomalous readings of these metrics are recorded at a certain distance from the origin of the configuration space where some widely used initialization schemes such as Xavier \citep{glorot} and Kaiming \citep{kaiming} are found. Thus, the Goldilocks zone is believed to be a hollow centered spherical shell that contains a high density of suitable initial points. Indeed, an appropriate parameter norm is crucial to avoid exploding and vanishing signals, and it seems reasonable that initializations of extreme norm (outside the Goldilocks zone) might suffer from this notorious issue. However, internal covariate shift is largely solved in practice by BatchNorm \citep{batchnorm}, and it can be directly accounted for in some special cases. For example, given an $L$-layer homogeneous network $f_{\alpha\theta}$ with an $\alpha$-scaled initialization \citep{dinh}, the appropriate logit variance and gradient norm can be restored by applying a carefully selected softmax temperature $T=\alpha^L$ and learning rate $\eta=\mathcal{O}(\alpha^{2})$. These adjustments ensure that $f_{\alpha\theta}$ follows exactly the same training trajectory and has the same initial excess of positive curvature as $f_{\theta}$ (Figure \ref{Fig:IntroGoldilocks}). Thus, contrary to the original claims by \citet{goldilocks}, the Goldilocks zone cannot be characterized by the initialization norm alone and is not even a subset of the configuration space. Instead, we find a more fundamental condition governing the excess of positive curvature of the loss in Section \ref{Sec:Goldilocks}.

Previous works also attempted to describe the Goldilocks zone more formally. \citet{emergent} develop an abstraction, the random logit model, in which they show that prevalence of positive curvature of cross-entropy loss vanishes with increased logit variance. \citet{eos} attribute the decrease in cross-entropy sharpness during optimization to the collapse of the softmax distribution, which also has implications for the Goldilocks zone. However, the phenomenon of excessive positive curvature---its relation to the initialization norm and trainability---remains unexplained. In this work, focusing on homogeneous networks, we derive a comprehensive description of the Goldilocks zone from the Gauss-Newton decomposition (Section \ref{Sec:Goldilocks}) of the cross-entropy loss Hessian and formally associate excess of positive curvature with certain properties of the network (Section \ref{Sec:Features}). In particular, we show that near-zero curvature results from a relatively low spectral norm of the G-term in the Hessian decomposition due to either saturated softmax or vanishing logit gradients, which naturally arise for high- and low-norm initializations, respectively. Inside the Goldilocks zone, we prove that the highest excess of positive curvature is observed for networks with low confidence, which in turn is associated with low initial loss and, for balanced datasets, with vanishing expected loss gradient.

When the conditions governing positive curvature of the loss are established, we inquire about their relation to model trainability and optimization dynamics. \citet{goldilocks} find that optimization on random subspaces of the parameter space converges only if they intersect the Goldilocks zone. For unconstrained optimization, \citet{gurari} discovered that gradients are mainly confined to a low-rank subspace spanned by the Hessian top eigenvectors and vanish along flatter directions. Since the excess of positive curvature manifests a larger separation between bulk and outlier eigenvalues, it should then be associated with a more robust top-eigenspace and, intuitively, a more informative training signal. Motivated to unveil this relationship, we use gradient descent to optimize homogeneous networks using a wide spectrum of initialization norms and learning rates and taxonomize the emergent behaviors. Interestingly, we find that successful training is not necessarily well aligned with the Goldilocks zone. We demonstrate setups where the slightest increase in the initialization norm of LeNet-5 leads to degenerate learning despite happening well within boundaries of the Goldilocks zone. These dynamics are marked by an increasing amount of zero logits and, to the best of our knowledge, we are the first to report this behavior.

\paragraph{Contributions.} This paper conducts an extensive study of the Goldilocks zone of homogeneous neural networks, both analytically and empirically. Our code is available at  \href{https://github.com/avysogorets/goldilocks-zone}{https://GitHub.com/avysogorets/goldilocks-zone}.
\begin{itemize}
\item In Section \ref{Sec:Goldilocks}, we demonstrate that the Goldilocks zone is not characterized by the initialization norm alone, refining prior beliefs of \citet{goldilocks}. Instead, we derive a more fundamental condition resulting in excess of positive curvature and find that it disappears due to saturated softmax on one end and vanishing logit gradients on the other.
\item In Section \ref{Sec:Features}, we closely study the interior of the Goldilocks zone and analytically associate excess of positive curvature with low model confidence, low initial loss, and low cross-entropy gradient norm.
\item In Section \ref{Sec:Trainability}, we report the evolution and performance of scaled homogeneous networks when optimized by gradient descent both inside and outside the Goldilocks zone. Our investigation shows that excess of positive curvature is an imperfect estimator of the initialization trainability and exhibits a range of interesting effects for initializations on the edge.
\end{itemize}

\section{Preliminaries \& Notation}
\label{Sec:Notation}
We begin by introducing the technical scope of this study, notation, and the essential background. We consider a standard $K$-way classification problem $\mathcal{D}=\{(X^{\mu}, y^{\mu})\}_{\mu=1}^N$ with targets $y^{\mu}\in[K]$. A neural network $f_{\theta}$, parameterized by a vector $\theta\in\mathbb{R}^P$, computes $K$ logits $\{z_k\}_{k=1}^K$ associated to a probability distribution $p$ via the softmax function $\sigma_T$ with a temperature parameter $T$:
\begin{align}\label{eq:p-logits}
p_k=[\sigma_T(z)]_k = \frac{\text{exp}[\frac{1}{T}z_k]}{\sum_{c=1}^K\text{exp}[\frac{1}{T}z_c]}\,.
\end{align}
By default, $T=1$, in which case we refer to softmax simply as $\sigma$. The corresponding cross-entropy loss is $\ell(p, y)=-\log p_y$. The Hessian matrix $H\in\mathbb{R}^{P\times P}$ holds the second derivatives of the loss at $\theta$: $H_{ij} = \partial^2\ell/\partial\theta_i\partial\theta_j$. In principle, all of the above quantities depend on one particular or a batch of inputs $(X^{\mu}, y^{\mu})$; we use the superscript $\mu$ to make this dependence explicit where needed. Since we are often concerned with the magnitude of network outputs at different parameter scales, we assume the inputs are standardized and bounded.

\paragraph{Homogeneous networks.} We restrict our analysis to homogeneous models satisfying $f_{\alpha\theta} = \alpha^{L}f_{\theta}$ for any scalar $\alpha>0$ where $L$ is the number of layers in $f$. This is a rather technical assumption: homogeneous models are widely used in practice and include ReLU networks without biases. \citet{goldilocks} focus on initially homogeneous models by setting biases to zero. Inhomogeneous networks encompass a wide range of architectures and design choices that require case-by-case analysis depending on the source of inhomogeneity (e.g., non-zero biases, residual connections, inhomogeneous activation functions, etc.). Many inhomogeneous cases may be quite degenerate in the context of initialization scaling. For example, upscaling the initialization of TanH-networks simply saturates the activations, blocking any signal propagation. Downscaling these networks sufficiently hard makes them into linear models. For another example, consider ReLU networks with non-zero biases and at least one hidden layer, which are inhomogeneous due to the different scales of biases and activations in hidden layers. When $\alpha\ll 1$, the output signal is dominated by biases; when $\alpha\gg1$ biases vanish in magnitude compared to the corresponding activations, making the model almost homogeneous. In Section \ref{Sec:Discussion}, we revisit inhomogeneous models and suggest how our analyses can be extended to these architectures as well.

\paragraph{Excess of positive curvature.}
In their work, \citet{goldilocks} define the Goldilocks zone rather informally as a region of excessive positive curvature of the loss, which can be unmistakably identified in Figure \ref{Fig:IntroGoldilocks}.  of positive curvature of the loss is defined as the excess of positive eigenvalues $\{\lambda_i\}_{i=1}^P$ of a loss Hessian $H$:
\begin{align}
\label{Eq:positive-curvature}
\frac{\text{Tr}(H)}{\lVert H\rVert_F} = \frac{\sum_{i=1}^P\lambda_i}{\sqrt{\sum_{i=1}^P\lambda_i^2}}.
\end{align}
Throughout the remainder of the paper, we refer to this quantity simply as \emph{positive curvature} for fluid presentation. Another metric used by \citet{goldilocks} is local convexity of the loss, which refers to the fraction of positive eigenvalues of $H$. These two metrics are intimately related and can detect the Goldilocks zone independently of each other, so we focus only on positive curvature throughout this paper. Since computing the full Hessian is intractable for most modern architectures, \citet{goldilocks} substitute it with a projection $H_d = R^{T}HR$ onto a low-rank random subspace with basis defined by a sparse matrix $R\in\mathbb{R}^{P\times d}$ with orthogonal columns. Given an initialization $\theta$, this amounts to computing the Hessian of the training loss of the model parameterized by $R\hat{\theta}+\theta$ with respect to latent $d$-dimensional parameters $\hat{\theta}$ at the origin of the chosen low-rank space, which is much more accessible. We adopt the same strategy and assume that the first- and second-order derivatives with respect to $\theta$ more generally represent derivatives with respect to \emph{trainable} $d$-dimensional parameters, which can be latent parameters $\hat{\theta}$ or the original model weights if we let $d=P$ and $R=I_P$. This technical nuisance affects none of our analyses, but we add further comments as it becomes necessary. We verify the validity of this practical approach by comparing it to Hutchinson's stochastic trace estimation in Appendix \ref{App:Hutchinson} \citep{hutch}.

\section{Revisiting the Goldilocks Zone}
\label{Sec:Goldilocks}
\citet{goldilocks} introduced the Goldilocks zone as a region of the parameter space with an excess of positive curvature and local convexity of the loss function, as defined in Section \ref{Sec:Notation}. Starting from a gold standard Kaiming initialization $\theta_0$, \citet{goldilocks} record these statistics over a ray $\{\alpha\theta_0\colon\alpha>0\}$ and read unusually high values when $\alpha$ falls within a relatively narrow range centered around $\alpha=1$ (Figure \ref{Fig:IntroGoldilocks} left). Moreover, they find that SGD constrained to a random subspace is successful only if it intersects the Goldilocks zone. Based on these observations, the authors suggest that the Goldilocks zone is a thick, hollow spherical shell about the origin in the configuration space, which is densely populated with initial points amenable for training.

A simple example shows that, strictly speaking, this visually appealing representation needs revision. For homogeneous models, the $\alpha$-scale transformation $\theta\rightarrow\alpha\theta$ with $\alpha>0$ effectively scales the underlying configuration space by $\alpha$ \citep{dinh}. We shall derive next that the cross-entropy loss landscape scales together with the configuration space when the softmax temperature satisfies $T=\alpha^{L}$, restoring positive curvature of the loss of the scaled model $f_{\alpha\theta}$ to that of the original model $f_{\theta}$ (see the middle plot in Figure \ref{Fig:IntroGoldilocks}). Thus, we argue that initialization norm has a coincidental relationship to the Goldilocks zone, calling for a refined, analytically driven characterization.

\paragraph{Gradients and Hessian of scaled models.}
We begin by simplifying the notation for clarity of presentation. Denote a scaled model $f_{\alpha\theta}$ by $f'$ and adopt a similar notation for all of its attributes ($\theta'=\alpha\theta$, etc.). The chain rule allows us to express gradients of the cross-entropy loss as
\begin{align}
\label{Eq:gradients}
\frac{\partial\ell}{\partial\theta}=\frac{1}{T}\left[-\frac{\partial z_y}{\partial \theta}+\sum_{k=1}^K\sigma_T(z_k)\frac{\partial z_k}{\partial \theta}\right].
\end{align}
By virtue of homogeneity, the logit gradients of the scaled model $f'$ satisfy $\partial z_k'/\partial \theta'=\alpha^{L-1}\partial z_k/\partial\theta$. For the special case of $T=\alpha^L$, we have $\sigma_{\alpha^L}(z'_k)=\sigma(z_k)$, giving $\nabla_{\theta'}\ell={\alpha}^{-1}\nabla_{\theta}{\ell}$. This tells us that the $\alpha$-scaled model $f'$ follows exactly the same optimization trajectory as $f$ if the ratio of their respective learning rates is $\alpha^2$. This factor ensures equal update norms relative to parameter norms across the two models. To derive a similar relationship for the Hessians of $f'$ and $f$, we turn to the Gauss-Newton decomposition \citep{sagun, emergent, papyan}. For a single training sample, we have:
\begin{align}
\label{Eq:GN}
H_{ij}=\underbrace{\sum_{k=1}^K\sum_{c=1}^K\frac{\partial z_k}{\partial\theta_i}\left[\nabla^2_{z}\ell\right]_{kc}\frac{\partial z_c}{\partial\theta_j}}_{\text{G-term } (\mathcal{G})}
+\underbrace{\sum_{k=1}^K\left[\nabla_{z}\ell\right]_k\frac{\partial^2z_k}{\partial\theta_j\partial\theta_i}}_{\text{H-term }(\mathcal{H})}\,,
\end{align}
where $\nabla^2_z\ell=\text{diag}(p)-pp^{\top}$ is the Hessian of the loss with respect to the logits and $\nabla_z\ell=p-\text{OH}(y)$ where $p$ is the softmax output and $\text{OH}(y)$ is a one-hot encoded tagret \citep{eos}. To emphasize the dependence of the G-term and H-term on $p$ we sometimes refer to them as $\mathcal{G}(p)$ and $\mathcal{H}(p)$, respectively. Note that $\mathcal{G}(p)=J^{\top}[\text{diag}(p)-pp^{\top}]J$ where $J$ is the Jacobian matrix. For the $\alpha$-scaled model $f'$ used together with softmax $\sigma_T$, we similarly get $\nabla^2_{z'}\ell'=T^{-2}\left[\text{diag}(p')-p'p'^{\top}\right]$, $\nabla_{z'}\ell'=T^{-1}(p'-\text{OH}(y))$. Combining this with homogeneity of gradients, we derive the Gauss-Newton decomposition of the Hessian of $f'$:
\begin{align}
\label{Eq:GNS}
H'&=\mathcal{G}'+\mathcal{H}'=\frac{\alpha^{2L-2}}{T^2}\mathcal{G}(p')+\frac{\alpha^{L-2}}{T}\mathcal{H}(p').
\end{align}
Linearity of differentiation ensures that this equation is valid for the full-batch Hessian, too. Indeed, the full G-term and the full H-term are just the averages of the per-sample quantities $\mathcal{G}^{\mu}$ and $\mathcal{H}^{\mu}$ defined in \cref{Eq:GN}, respectively. Thus, we will abuse this notation and refer to their full counterparts in the same way where appropriate. In this case, the probability vectors $p$ and $p'$ can be viewed as matrices.

Returning to our discussion on the shape of the Goldilocks zone, we remark that $T=\alpha^L$ yields $H'=\alpha^{-2}H$ (as this temperature ensures $p=p'$). Since the measures of positive curvature and local convexity are robust to scaling of the Hessian matrix, we conclude that \emph{initialization of any norm can be in the Goldilocks zone provided an appropriate softmax temperature} (see Figure \ref{Fig:IntroGoldilocks}). Now that the connection between the initialization norm and the Goldilocks zone is much less credible, we leverage \cref{Eq:GNS} to establish the fundamental principles governing positive curvature in neural networks.

\paragraph{Gauss-Newton decomposition.}
The Gauss-Newton decomposition in \cref{Eq:GN} is a common entry point for many studies on the Hessian of large neural networks. The Hessian exhibits a ``bulk-outlier'' eigenspectrum with the majority of eigenvalues small and clustered around zero and only a handful of large positive outliers \citep{sagun, gurari, ghorbani}. This decomposition is inherited from the individual spectra of $\mathcal{G}$ and $\mathcal{H}$ with the top and the bulk eigenvalues attributed to these two terms, respectively. We present a survey of works concerned with this phenomenon in Appendix \ref{App:GaussNewton}.

\begin{figure}[ht]
\centering
\includegraphics[width=0.5\linewidth]{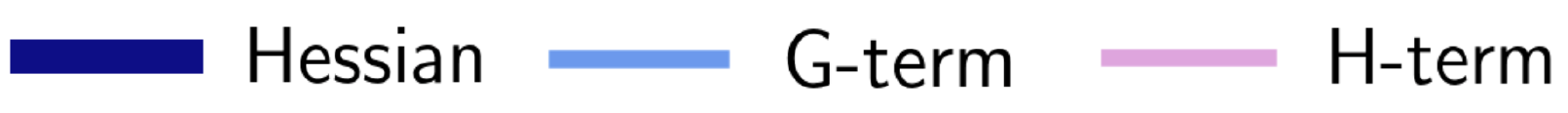}
\includegraphics[width=\linewidth]{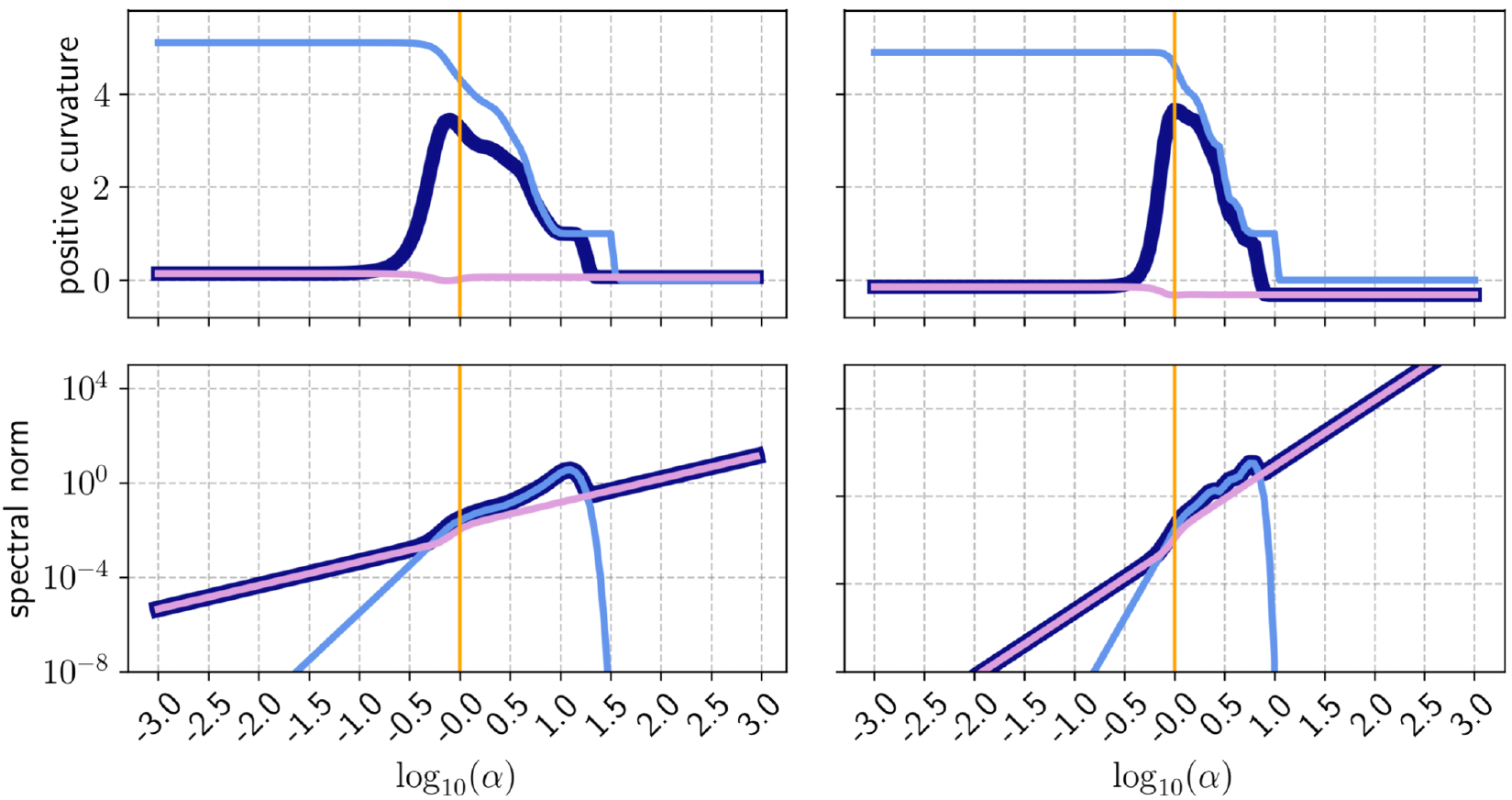}
\caption{Positive curvature (top) and spectral norm (bottom) of the Hessian, G-term, and H-term across initialization scales. We computed these quantities on a low-rank subspace with $d=50$. \textbf{Left:} LeNet-300-100 (fully-connected) on FashionMNIST; \textbf{Right:} LeNet-5 (convolutional) on CIFAR-10.}
\label{Fig:TrueGoldilocks}
\end{figure}

\paragraph{Rediscovering the Goldilocks zone.} We are now in position to describe the exact conditions that result in prevalence of positive curvature---the hallmark of the Goldilocks zone. The eigenstructures of the individual terms in the Gauss-Newton decomposition of the Hessian suggest that the G-term is the one and only source of positive curvature of the loss, and so 
\begin{align}
\label{Eq:Condition}
\lVert \mathcal{G}'\rVert_2\gtrsim\lVert \mathcal{H}'\rVert_2
\end{align}
is a necessary and sufficient condition for the Goldilocks zone. We observe this exact correspondence in Figure \ref{Fig:TrueGoldilocks}. Otherwise, when $\lVert \mathcal{G}'\rVert_2$ is sufficiently small relative to $\lVert \mathcal{H}'\rVert_2$, the bulk-like eigenspectrum of the H-term swallows the outliers of the G-term and is inherited by the Hessian, resulting in a near-zero positive curvature. This sudden change in the eigenstructure is known as the BBP phase transition \citep{bbp, emergent}. Normally (when $\alpha=1$ and $T=1$), we expect the above inequality to be true. Figure \ref{Fig:TrueGoldilocks} (bottom) confirms that the G-term has an edge over the H-term within a neighborhood around the unaltered model (orange bar), which also corresponds to the Goldilocks zone.

Therefore, it is left to understand when and why \cref{Eq:Condition} is violated. Figure \ref{Fig:TrueGoldilocks} reveals that this occurs at sufficiently low or sufficiently high values of $\alpha$ given a fixed temperature $T=1$. For $\alpha>1$, the norm of the G-term plummets to zero when the logit variance becomes sufficiently large for the softmax output $p'$ to collapse to a one-hot distribution for all training samples. In this scenario, $\text{diag}(p')-p'p'^{\top}$ is identically zero, as has been observed by \citet{eos} in the context of progressive sharpening. For us, this implies that $\mathcal{G}'(p')=J'^{\top}[\text{diag}(p')-p'p'^{\top}]J'$ is identically zero as well, making the Hessian equal to the H-term. \citet{emergent} observed the causal relationship between the increased logit variance and vanishing positive curvature but never explained it analytically.

For $\alpha<1$, on the other hand, the G-term does not completely vanish; in fact, it achieves the largest excess of positive eigenvalues at low initialization scales (see the top plots in Figure \ref{Fig:TrueGoldilocks}), suggesting that higher entropy predictions should generally be associated with larger positive curvature. According to \cref{Eq:GNS}, $\lVert \mathcal{G}'\rVert_2=\mathcal{O}(\alpha^{2L-2})$ while $\lVert \mathcal{H}'\rVert_2=\mathcal{O}(\alpha^{L-2})$. Thus, the G-term simply decays faster than the H-term as $\alpha\rightarrow 0$, eventually letting it dominate in the Gauss-Newton decomposition. The higher decay rate of the G-term comes from the cross-class product of logit gradients that vanish as $\mathcal{O}(\alpha^{L-1})$ each. Thus, we call this phenomenon \emph{vanishing logit gradient} and emphasize that vanishing \emph{loss} gradient can in fact co-occur with high positive curvature as we demonstrate in the next section.

\section{Features of the Goldilocks Zone}
\label{Sec:Features}
In this section, we inquire about the properties of the interior of the Goldilocks zone. In particular, we will associate extreme values of positive curvature with certain features of the network, initialization, and the data. To this end, we study the spectral properties of the G-term that, provided \cref{Eq:Condition} holds, transfer to the loss Hessian as well. \cref{Eq:GNS} reveals that $\mathcal{G}'(p)\propto \mathcal{G}(p)$ for the same $p$, so that structural changes in the eigenspectrum of the G-term at scaled initializations are due to the changing softmax output $p$ alone. Therefore, we can drop the superscript and study the behavior of $\mathcal{G}$ as a function of $p$, which is the goal of this section. In particular, we will find that positive curvature is associated with low model confidence, which, in turn, leads to vanishing expected loss gradients for balanced datasets and low initial loss.

At the basis of our analysis is the random model introduced by \citet{emergent}. They observed that same-logit gradients cluster across training samples, and that the $K$ corresponding logit gradient means $c_k$ are nearly orthogonal to each other. In light of this observation, they model $c_k\sim\mathcal{N}(0,\sigma_c^2I_P)$ and, for every input $X^{\mu}$, let the corresponding $k$-th logit gradient be $\nabla_{\theta}f_k(X^{\mu}) = c_k+e^{\mu}_k$ with iid residuals $e_{k}^{\mu}\sim\mathcal{N}(0,\sigma_e^2I_P)$. Recall from Section \ref{Sec:Notation} that, in practice, we compute positive curvature by projecting the Hessian onto a low-rank hyperplane given by an orthonormal basis $R\in\mathbb{R}^{P\times d}$, so we actually care about gradients with respect to the $d$-dimensional trainable parameters. Conveniently, the random model still holds in this low-rank subspace since the corresponding gradients are just $R^{\top}c_k+R^{\top}e_k^{\mu}$ where $R^{\top}c_k\sim\mathcal{N}(0,\sigma_c^2I_d)$ and $R^{\top}e^{\mu}_k\sim\mathcal{N}(0,\sigma_e^2I_d)$ as linear transformations of isotropic Gaussian variables by $R^{\top}$ satisfying $R^{\top}R = I_d$. Thus, we drop the transformation and simply assume $c_k\sim\mathcal{N}(0,\sigma_c^2I_d)$ and $e^{\mu}_k\sim\mathcal{N}(0,\sigma_e^2I_d)$. Note that, in particular, the variance of logit gradients and the corresponding residuals does not depend on $d$. We make two more simplifying assumptions and require $c_k$ to be pairwise orthogonal and have equal squared length $\lVert c_k\rVert^2 = \mathbb{E}\lVert c_k\rVert^2 = d\sigma^2_c$.

\begin{figure}[t]
\centering
\includegraphics[width=\linewidth]{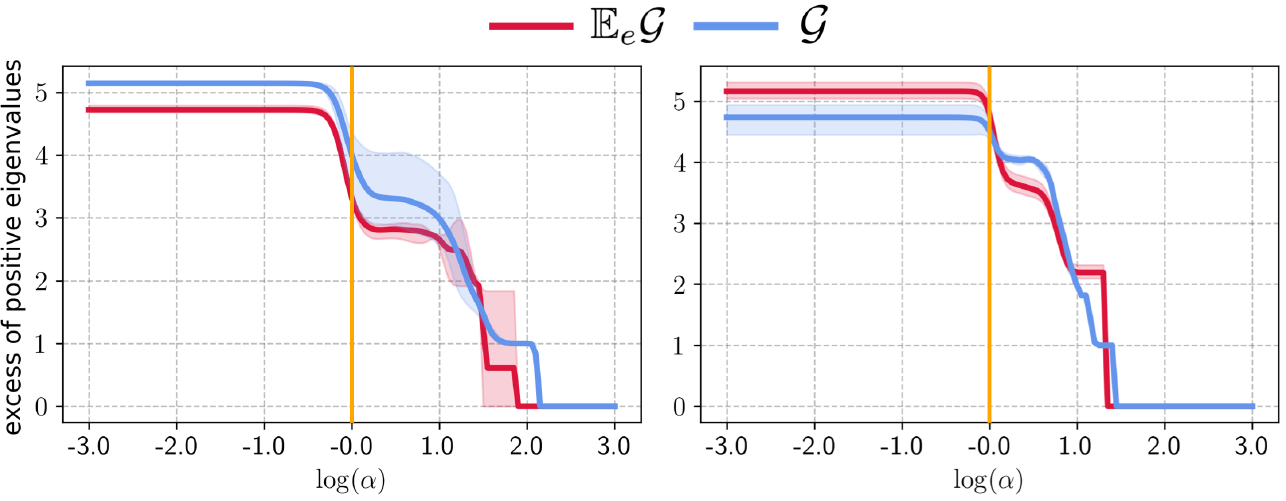}
\caption{Excess of positive eigenvalues of the true G-term $\mathcal{G}$ and the expected G-term $\mathbb{E}\mathcal{G}$ from \cref{Eq:trace-over-norm} computed using logits associated with different initialization scales (model confidence). We used a low-rank subspace with $d=50$. Error bands represent min/max across $3$ seeds. \textbf{Left:} LeNet-300-100 on FashionMNIST; \textbf{Right:} LeNet-5 on CIFAR-10.}
\label{Fig:ExpectedG}
\end{figure}

For a single training sample, let $C$ and $E$ be $k\times d$ matrices of logit gradient means and the corresponding residuals (i.e., $C_k = c_k$ and $E_k = e_k$), respectively; the Jacobian is then $J=C+E$ and so $\mathcal{G} = (C+E)^{\top}[\text{diag}(p)-pp^{\top}](C+E)$. Taking expectation over data, we get
\begin{align}
\label{Eq:EG}
\mathbb{E}_e\mathcal{G} &= C^{\top}[\text{diag}(p)-pp^{\top}]C+(1-\lVert p\rVert^2)\sigma_e^2I_d\nonumber\\
&\equiv d\sigma_c^2\begin{bmatrix}\text{diag}(p)-pp^{\top} & 0 \\ 0 & 0\end{bmatrix} + (1-\lVert p\rVert^2)\sigma_e^2I_d
\end{align}

\begin{figure}[b]
\centering
\includegraphics[width=\linewidth]{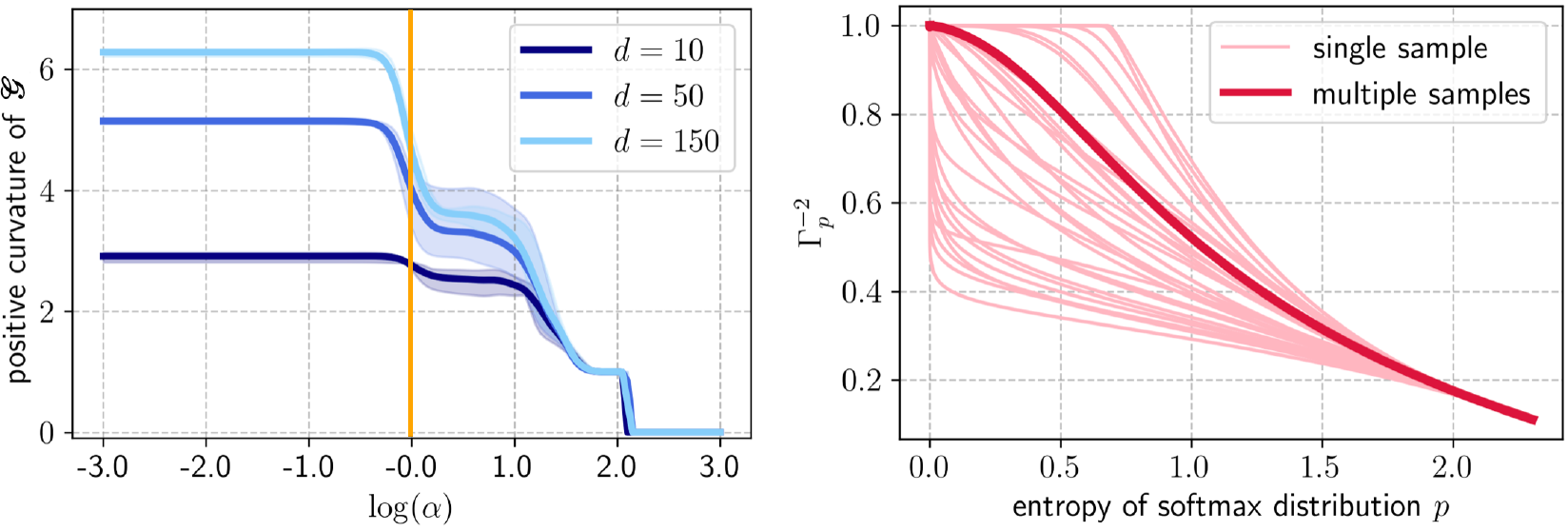}
\caption{\textbf{Left:} The dependence of excess of positive eigenvalues on the dimension $d$ of the low-rank hyperplane used to compute the projected G-term. \textbf{Right:} $\Gamma_p^{-2}$ of the matrix $\text{diag}(p)-pp^{\top}$ where vectors $p$ are produced by scaling $30$ different logit sets by $\alpha\in[10^{-2}, 10^2]$ ($30$ pink curves). The red curve corresponds to $\Gamma_p^{-2}$ computed for the average matrix $\frac{1}{30}\sum_{\mu=1}^{30}\text{diag}(p^{\mu})-p^{\mu}{p^{\mu}}^{\top}$.}
\label{Fig:Analysis}
\end{figure}

\begin{figure*}[t]
\centering
\subfloat[\centering] {{\includegraphics[width=0.7\textwidth]{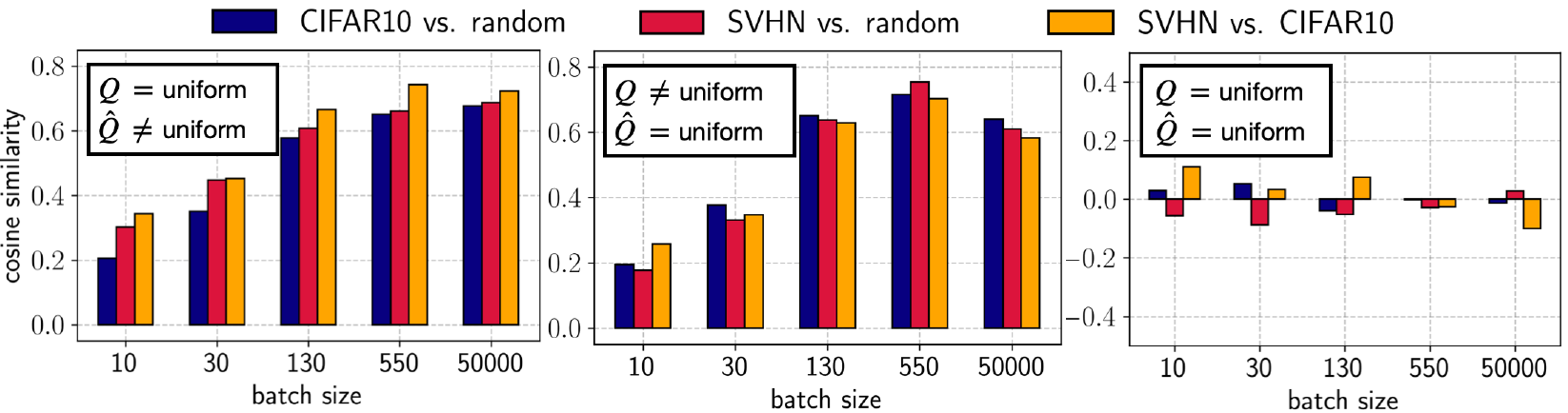}}}%
\quad
\subfloat[\centering]{{\includegraphics[width=0.245\textwidth]{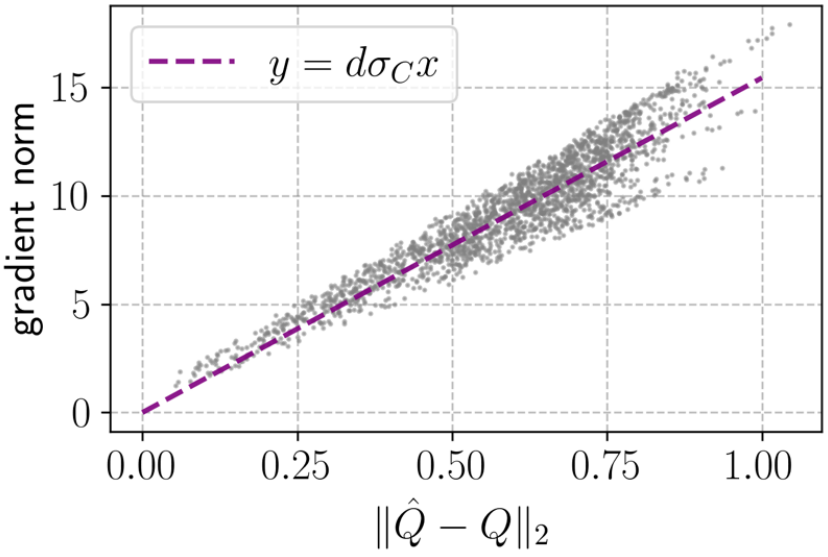}}}%
 \captionsetup{font=small}
 \vspace{-10px}
\caption{The effects of the average softmax output $\hat{Q}$ and the actual target prior $Q$ on the batch gradient. \textbf{(a)}: Cosine similarity of gradients computed on different datasets (SVHN, CIFAR-10, and a randomly generated dataset) at the same initialization of LeNet-5 (downscaled by $\alpha=0.01$ when $\hat{Q}=\text{uniform}$). To achieve $Q\ne \text{uniform}$, we inject artificial class imbalance by subsampling datasets using a procedure suggested by \citet{cui}. \textbf{(b):} Given a fixed unscaled initialization of LeNet-5, we sample $2,000$ different class priors $Q$ uniformly from a probability simplex $\Delta_{10}$. For each of them, we sample a subset of CIFAR-10 with $5,000$, compute the cross-entropy gradient, and plot its norm against $\lVert\hat{Q}-Q\rVert_2$. As we predicted, the norms follow a linear trend with the predicted slope ($\sigma_c^2$ was estimated on the entire CIFAR-10 dataset; $d=P=61,170$).}%
\label{Fig:GradOverlap}
\end{figure*}

where in the second line, we switch to the orthonormal basis of the normalized logit gradient means $c_k$. We can now compute the excess of positive eigenvalues of $\mathbb{E}_e\mathcal{G}$ directly from its elements. Assuming that $p$ is not one-hot, we get
\begin{align}
\label{Eq:trace-over-norm}
\frac{\text{Tr}(\mathbb{E}_e\mathcal{G})}{\lVert \mathbb{E}_e\mathcal{G}\rVert_F} = \frac{\sqrt{d}(\sigma_c^2+\sigma_e^2)}{\sqrt{\sigma_e^4+2\sigma_e^2\sigma_c^2+d\sigma_c^4{\Gamma}_p^{-2}}}
\end{align}
where $\Gamma_p$ is the excess of positive eigenvalues of $\text{diag}(p)-pp^{\top}$ as defined in \cref{Eq:positive-curvature}. For the expected G-term computed on a batch of samples with potentially different softmax probabilities $p^{\mu}$, $\Gamma_p$ computes excess of positive eigenvalues of the average matrix $\frac{1}{B}\sum_{\mu=1}^B\text{diag}(p^{\mu})-p^{\mu}{p^{\mu}}^{\top}$. Having empirically estimated values $\sigma_c^2$ and $\sigma_e^2$, we use \cref{Eq:trace-over-norm} to compute the excess of positive eigenvalues of $\mathbb{E}_e\mathcal{G}$ on real softmax outputs $p$ found by scaling initialization, and find that it approximates the excess of positive eigenvalues of the true G-term across varying model confidence levels very well (Figure \ref{Fig:ExpectedG}).

\paragraph{Positive curvature \& dimension $d$.} We can now explain the dependence of positive curvature on the dimension $d$ of the hyperplane used to project the model parameters, which was first noticed by \citet{goldilocks}, see the left plot in Figure \ref{Fig:Analysis}. Indeed, according to \cref{Eq:trace-over-norm}, larger $d$ are associated with higher excess of positive eigenvalues of the expected G-term given the same softmax outputs $p$.

\paragraph{Positive curvature \& model confidence.}
Figure \ref{Fig:TrueGoldilocks} shows that excess of positive eigenvalues of the G-term monotonically decreases, reaching zero at some $\alpha>1$ due to a collapsing softmax. Since larger $\alpha$ are associated with colder softmax distributions, it is natural to hypothesize that excess of positive eigenvalues of the G-term is inversely related to model confidence. \cref{Eq:trace-over-norm} reveals that excess of positive eigenvalues of the expected G-term is directly related to that of $\text{diag}(p)-pp^{\top}$ and is larger for smaller values of $\Gamma_p^{-2}$. Despite the seemingly nice form of this matrix, $\Gamma_p$ is difficult to analyze algebraically; still, a few general remarks shed light on the above relationship. First, \cref{Eq:EG} can be used to show that $\mathbb{E}\mathcal{G}$ has a ``bi-level'' eigenspectrum that yields maximal excess of positive eigenvalues when the softmax output $p$ is uniform (Appendix~\ref{App:EigenvaluesEG}). Second, as seen from \cref{Eq:positive-curvature}, excess of positive eigenvalues of any rank-$r$ positive semidefinite matrix is just the ratio of L1 and L2 norms of its $r$ non-zero eigenvalues, which is known to fall between $1$ and $\sqrt{r}$. Therefore, $\Gamma_p\in[1, \sqrt{K-1}]$, and it achieves its maximal value when $p$ is uniform. In Appendix \ref{App:Limit}, we prove that $\Gamma_p\rightarrow 1$ as $p$ collapses to a one-hot vector due to the increasing initialization scale $\alpha$. Moreover, Figure \ref{Fig:Analysis} (right) reveals that $\Gamma_p$ is, in fact, monotonic between its extreme values with respect to the entropy of $p$. Overall, this suggests that $\Gamma^{-2}_p$ in \cref{Eq:trace-over-norm} grows with model confidence, which supports our hypothesis.

\paragraph{Model confidence \& vanishing gradients.} In Section \ref{Sec:Goldilocks}, we saw that vanishing logit gradients associated with small-norm network initializations diminish the role of the G-term in the Gauss-Newton decomposition, leading to zero positive curvature of the loss. The term ``vanishing gradients'' commonly refers to the condition of signal decay during backpropagation to deeper layers, emerging either due to saturated activation functions or small spectral radii of parameter matrices \citep{vanishing1, vanishing2}. The vanishing logit gradients observed at low initialization scales of homogeneous networks fall in the latter category. Using the random model of \citet{emergent}, we derive yet another, previously unknown type of vanishing cross-entropy gradients caused by the match between the average softmax output of the model $\hat{Q}=\mathbb{E}p \approx\frac{1}{N}\sum_{\mu}p^{\mu}$ and the dataset prior $Q =\mathbb{E}\text{OH}(y)\approx\frac{1}{N}\sum_{\mu}\text{OH}(y^{\mu})$ where expectation is taken over the data distribution. To this end, consider the expected batch cross-entropy loss gradient:
\begin{align}
\mathbb{E}_E\left[\frac{\partial\ell}{\partial\theta}\right] &= \mathbb{E}_E\left[\frac{1}{B}\sum_{\mu=1}^B\sum_{k=1}^K(c_k+e_{k}^{\mu})(p_k^{\mu}-\mathbf{1}_{y=k})\right]\nonumber\\
&=\sum_{k=1}^Kc_k(\hat{Q}_k-Q_k)
\end{align}
By the orthogonality assumption on logit gradient means $c_k$, the norm of this quantity is $d\sigma_c^2\lVert \hat{Q}-Q\rVert$. In Figure \ref{Fig:GradOverlap}b, we confirm this linear relationship on LeNet-5 shockingly well. Normally, randomly-initialized models are biased and have non-uniform average priors $\hat{Q}$, so that $\lVert\hat{Q}-Q\rVert>0$. In this case, we expect loss gradients computed on two potentially unrelated but balanced datasets to have considerable overlap as both should align with the non-zero expected gradient. Indeed, Figure \ref{Fig:GradOverlap}a shows that gradients of LeNet-5 computed on CIFAR-10 and on random images drawn from a standard Gaussian have cosine similarity $>0.7$ when priors $Q$ and $\hat{Q}$ disagree, which is quite noteworthy for a 61,170-dimensional space. In contrast, for low-confident models that always predict uniform distribution (e.g., when $\alpha\ll 1$, see Figure \ref{Fig:GradOverlap}a right), logit gradients cancel each other out when computed on a large balanced batch. Hence, rather counterintuitively, we derive that large positive curvature is associated with low-norm cross-entropy gradients for balanced datasets. 

\begin{figure}[h]
\centering
\includegraphics[width=0.95\linewidth]{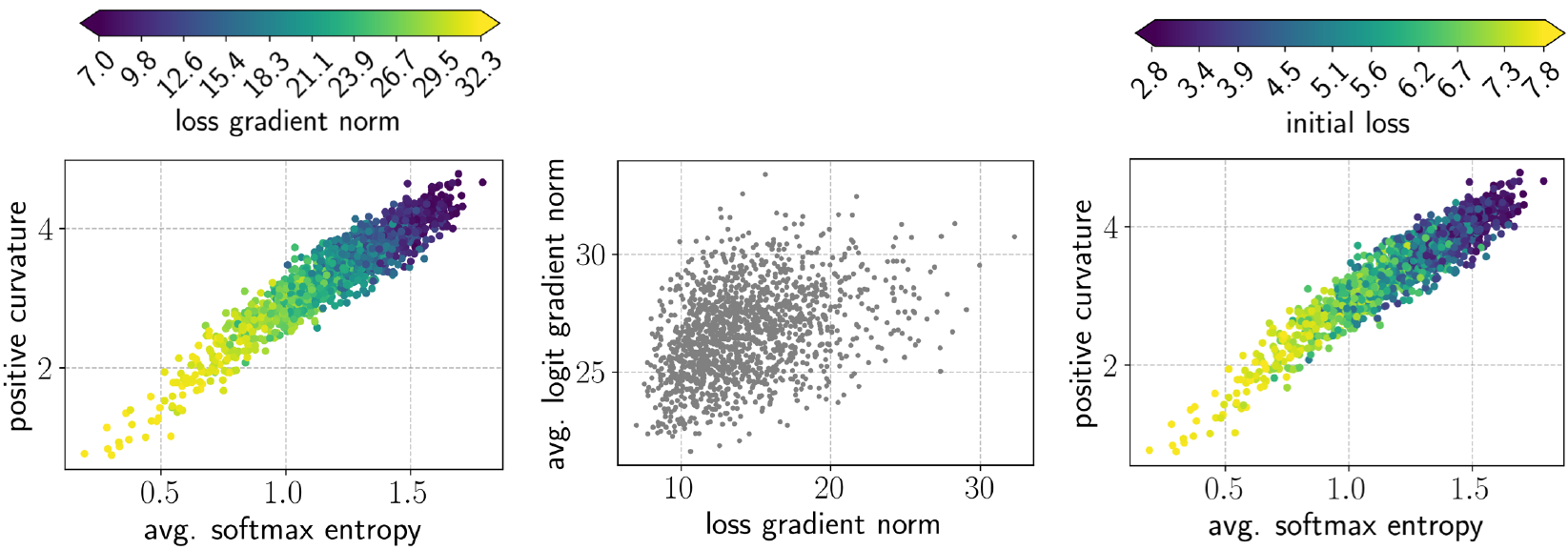}
\caption{The interplay between model confidence (avg. softmax entropy), loss gradient norm, positive curvature, and the initial loss value. Given a single balanced batch of $3,000$ CIFAR-10 images, we sampled $2,000$ Kaiming initializations for LeNet-5 and computed the above statistics for each of them.}
\label{Fig:Scatters}
\end{figure}

\paragraph{Model confidence \& initial loss.} \citet{goldilocks} observed that initializations with higher positive curvature tend to also have lower initial loss, which is often considered favorable for learning \citep{tempcheck}. Now, this relationship comes naturally since both phenomena are associated with low-confidence models. Indeed, the expected cross-entropy loss on samples from class $k$ is $\mathbb{E}_{p|y=k}(-\log p_{k})$, which is lower bounded by $-\log\mathbb{E}_{p|y=k}p_k$ by virtue of Jensen's inequality. If the softmax output $p$ is independent of the input's class identity $y$---a reasonable assumption for randomly-initialized models---then the lower bound is just $-\log\mathbb{E}p_k=-\log\hat{Q}_k$. Thus, the total expected loss is no less than $\sum_{k=1}^{K}(-\log\hat{Q}_k)\geq K\log K$ with equality when the expected model prediction $\hat{Q}$ is uniform. In principle, the model does not have to always be uncertain for $\hat{Q}$ to be uniform; the predictions can even be one-hot if all classes are equally-likely to be chosen. However, this symmetry is practically unachievable for randomly-initialized models, not to mention that the above bound becomes quite loose when the model makes confident mistakes. The right plot in Figure \ref{Fig:Scatters} confirms that lower initial losses correspond to higher average entropy of softmax output and, hence, larger positive curvature.

\paragraph{Summary.} In this Section, we scrutinized the interior of the Goldilocks zone by presenting novel observations and justifying some existing claims made in previous studies. In particular, we related high positive curvature to low model confidence, vanishing expected full-batch gradients for balanced datasets, and low initial loss, as is once again demonstrated in Figure \ref{Fig:Scatters}. 

\begin{figure}[h]
\centering
\includegraphics[width=\linewidth]{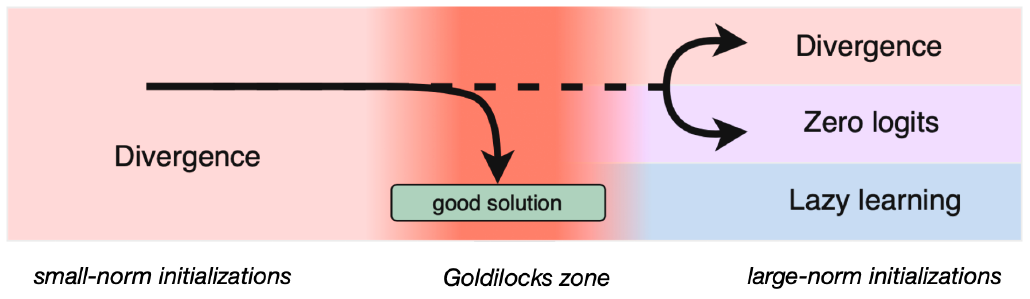}
\caption{Overview of the training outcomes observed in this study. Homogeneous networks with large-norm initializations either diverge, develop an increasing number of samples with zero logits, or learn in the lazy regime. Homogeneous networks with small-norm initializations diverge as long as their output remains uniform and either circle back to the Goldilocks zone and train normally, or continue diverging with a potential to develop zero logits and remain trapped in that regime.}
\label{Fig:Cartoon}
\end{figure}

\section{Connections to Optimization}
\label{Sec:Trainability}
Previous works suggest that, in a standard setup (i.e., within the Goldilocks zone), the top eigenspace of the training loss Hessian plays an important role during optimization. \citet{gurari} demonstrate that gradients are largely confined to that space, and \citet{ben-22} prove this phenomenon for two-layer neural networks. Outside the Goldilocks zone, where \cref{Eq:Condition} no longer holds, the loss curvature vanishes along all directions and so gradient descent must behave differently. In this section, we characterize the behavior of gradient descent outside the Goldilocks zone and identify some failure modes that result in poor convergence or generalization as shown in Figure \ref{Fig:Cartoon}.

\paragraph{Experimental setup.} We optimize $\alpha$-scaled homogeneous networks using vanilla full-batch gradient descent for 20,000 epochs. In particular, we use a fully-connected LeNet-300-100 with two hidden layers on FashionMNIST \citep{fmnist} and a convolutional LeNet-5 with $4$ hidden layers on CIFAR-10 \citep{lenets, cifar}, all implemented in PyTorch \citep{pytorch}. We set softmax temperature to $T=1$ to link the Goldilocks zone to the initialization norm. Recall from Section \ref{Sec:Goldilocks} that logit gradients of the $\alpha$-scaled network $f'$ are $\alpha^{L-1}$ times the respective logit gradients of the unscaled model $f$, so we multiply a preset learning rate $\eta_0$ by $\alpha^{2-L}$ to ensure that updates are initially commensurate to the weights of $f'$. In fact, this correction factor ensures that $f'$ has exactly the same training dynamics as $f$ provided that logit gradients are combined using the same softmax output $p=p'$ to compute the update vector as indicated in \cref{Eq:gradients}. From this perspective, our experiments essentially constitute an ablation study examining the effects of extreme model (un)confidence, which corresponds to the exterior of the Goldilocks zone in our setup.

\begin{figure}[h]
\centering
\includegraphics[width=\linewidth]{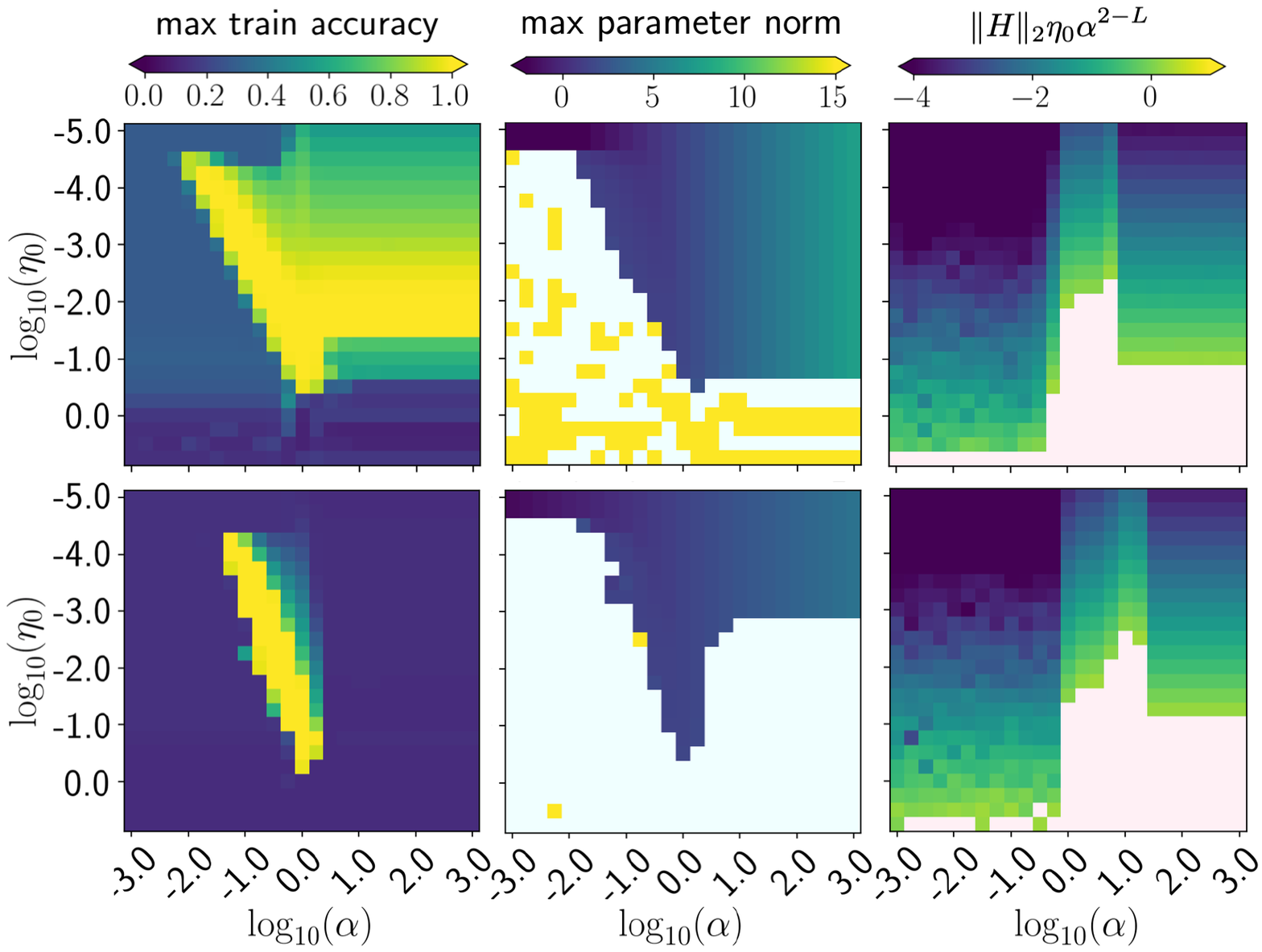}
\caption{Training statistics of LeNet-300-100 (top) and LeNet-5 (bottom) optimized with gradient descent across initialization scales $\alpha$ and base learning rates $\eta_0$. \textbf{Left:} maximum achieved training accuracy. Test accuracy exhibits the same general patterns and is omitted. \textbf{Middle:} maximum achieved parameter norm; azure values correspond to models that develop over 75\% zero logits and never recover from this regime. \textbf{Right:} product of the effective learning rate $\eta = \eta_0\alpha^{2-L}$ with the initial loss curvature $\lVert H\rVert_2$; inadmissible configurations with values above $2$ are shown in pink. The highlighted strip corresponds to the Goldilocks zone where the G-term is dominant (cf. Figure \ref{Fig:TrueGoldilocks}).}
\label{Fig:cmaps}
\end{figure}

\paragraph{Main results.} Our findings are summarized in Figure \ref{Fig:Cartoon}. As long as the softmax ouptut remains uniform ($\alpha<1$), homogeneous models remain in the divergence regime characterized by an increase of the parameter norm in accordance with previous studies \citep{omnigrok}. Once the weights become sufficiently large for the predictions to become non-uniform, the network circles back to the Goldilocks zone and trains normally thereafter provided that the learning rate is approximately admissible, i.e., $\eta=\eta_0\alpha^{2-L}<2/\lVert H\rVert_2$ \citep{admiss, eos, li22}. Hence, in Figure \ref{Fig:cmaps} (left) we observe a linear separation between trainable and divergent setups that extends well into the low-norm initialization region and way beyond the Goldilocks zone (see Appendix \ref{App:Extended} for further discussion). When the learning rate is too high, however, the network breezes past the Goldilocks zone and either diverges to infinity or develops an increasing amount of zero logits (middle column in Figure \ref{Fig:cmaps}). This happens as some negative weights grow much faster than the positive ones, promoting zero activations within the network by virtue of ReLU.

\begin{figure}[t]
\centering
\includegraphics[width=\linewidth]{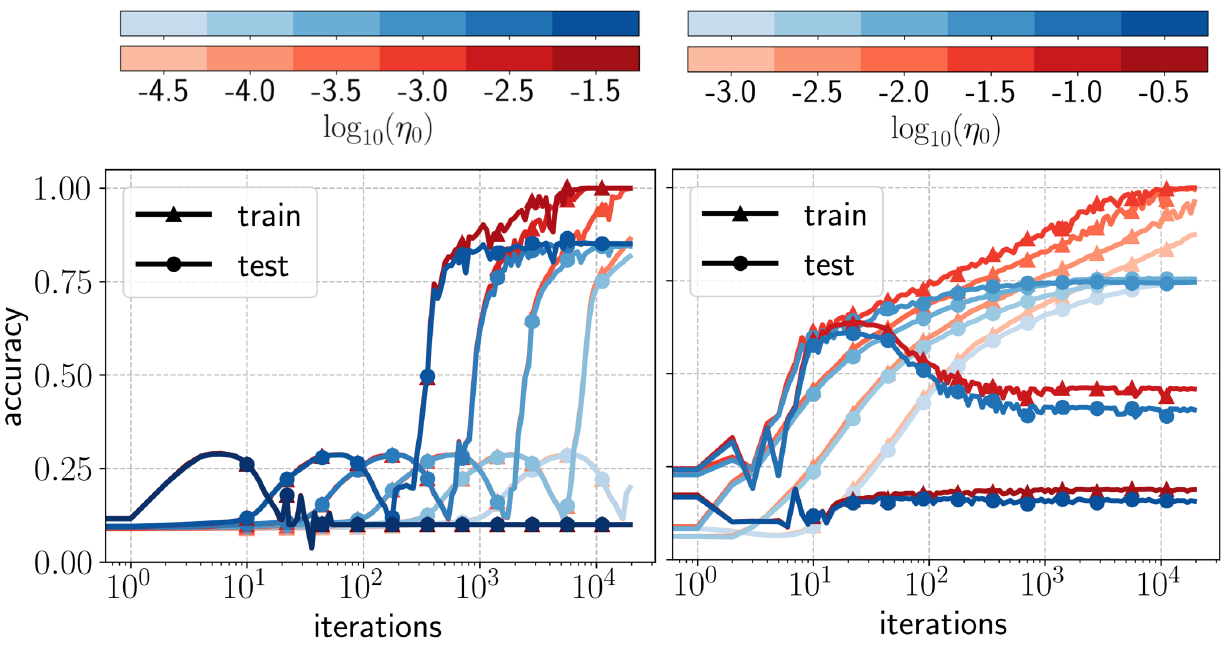}
\caption{Training curves of LeNet-300-100 optimized with full-batch gradient descent on FashionMNIST across different base learning rates $\eta_0$ (note that the effective learning rate is $\eta_0\alpha^{2-L}$). \textbf{Left:} low-confidence initialization with $\alpha=0.01$. \textbf{Right:} high-confidence initialization with $\alpha=1,000$.}
\label{Fig:Dynamics}
\end{figure}

Homogeneous networks with large-norm initializations ($\alpha>1$) exhibit a more diverse collection of behaviors. Previous works suggest that these models adhere to the lazy learning regime characterized by approximately linear optimization dynamics \citep{chizat, moroshko, new-grokking}. We confirm this observation for LeNet-300-100 but not for LeNet-5. Figure \ref{Fig:Dynamics} shows that, while the training error of LeNet-300-100 reaches zero, the test accuracy saturates at 75\%, which is 10\% lower than that of the unscaled network, suggesting lazy learning. In contrast, even the slightest increase in the initialization norm drives LeNet-5 to be completely untrainable (Figure \ref{Fig:cmaps} bottom-left), which is not well-aligned with the Goldilocks zone (Figure \ref{Fig:cmaps} bottom-right). In these cases, the network fails to train and develops zero logits (Figure \ref{Fig:cmaps} bottom-center). Unlike the case with $\alpha<1$, however, here zero logits often emerge when the weights are completely balanced, presenting a rather mysterious phenomenon.

\citet{omnigrok} and our own experiments show that networks with large initialization and no regularization neither diverge nor circle back to the Goldilocks zone. Thus, as long as these models are inaccurate, their loss remains extremely large and scales exponentially with $\alpha$ \citep{goldilocks}. Thus, we hypothesize that, unable to learn meaningful representations, gradient descent finetunes the weights of confident but inaccurate networks to make logits of misclassified samples zero in an attempt to reduce the exploding loss to just $\log(K)$---the value of cross-entropy under uniform softmax distribution. In favor of this intuition, we discover that models trained on random labels tend to develop zero logits as well. This condition arises early in training and disappears, provided that the labels remain fixed, as the network memorizes the correct output. On the other hand, it only gets worse if labels are randomized on every iteration, in which case the lowest possible loss is achieved by always making uniform predictions.

\paragraph{Goldilocks zone \& trainability.} The observations above reveal that homogeneous networks may not only converge when initialized outside the Goldilocks zone but also fail to learn when optimized well within its boundaries (Figure \ref{Fig:cmaps}). The former occurs as the model circles back to the Goldilocks zone ($\alpha\ll 1$) or when it learns the lazy solution ($\alpha\gg 1$), and the latter---as it develops zero logits. Indeed, LeNet-5 does not perform above random for any $\alpha>1.8$ and any learning rate, whereas the corresponding Goldilocks zone spans $\alpha\in[1, 18]$. Lastly, Figure \ref{Fig:cmaps} simply illustrates that phase transitions between different optimization regimes are not necessarily aligned with the Goldilocks zone, which is naturally highlighted in the right plots. Thus, we argue that positive curvature of the loss by itself is not a precise metric to evaluate the quality of initializations in terms of network trainability and generalization, and that further investigations are needed to clarify this connection.

\section{Conclusion}
\label{Sec:Discussion}
This paper studies the Goldilocks zone of neural network initializations in the context of homogeneous architectures, both analytically and empirically. We demonstrate that the excess of positive curvature of the training loss requires a robust top eigenspace of the Hessian inherited from its positive semi-definite component, the G-term and not a particular initialization norm per se. Normally, the G-term dominates in the Gauss-Newton decomposition of the Hessian but vanishes, for example, at extreme initialization scales or softmax temperatures, which is formally shown in \cref{Eq:GNS}. We study properties of the G-term itself and relate high positive curvature to low model confidence, low initial cross-entropy loss, and a previously unknown type of vanishing gradient caused by the match between class priors found in the data and computed by the model. Finally, we report the training behavior of homogeneous networks optimized by gradient descent from a wide variety of initialization norms and uncover interesting training modalities at the edges of the Goldilocks zone (Figure \ref{Fig:Cartoon}). Furthermore, we find that the connection between successful model training and the Goldilocks zone needs reconsideration. For example, homogeneous networks sometimes develop zero logits for an increasing number of inputs even when initialized and trained within the Goldilocks zone; thus, we argue against using positive curvature of the loss as a reliable estimate of the model performance.

\paragraph{Limitations \& future work.} The scope of our study is limited to homogeneous neural networks. At the same time, the methodology developed in our study is applicable to inhomogeneous models and does not require scaling of the initialization that could potentially introduce unexpected and difficult to analyze degeneracies to the model (e.g., saturation of the activation functions such as TanH or Sigmoid). Recall that our analyses in Section \ref{Sec:Goldilocks} indicates that the Goldilocks zone can be exactly replicated by adjusting the logit temperature $T$ instead of $\alpha$-scaling the initialization, see Figure \ref{Fig:IntroGoldilocks} and \cref{Eq:GNS}. Thus, in our study of the Goldilocks zone and positive curvature, $\alpha$-scaling of the initialization norm can be equivalently replaced by varying $T$. This does not imply that our conclusions for homogeneous architectures transfer to inhomogeneous ones. Instead, our findings in Section \ref{Sec:Goldilocks} allow for repeating the analyses and experiments regarding the Goldilocks zone presented in Sections \ref{Sec:Features} and \ref{Sec:Trainability} for any type of model without having to scale the initialization.

Even though we scope our study at the initialization stage to conform with previous research, much of our analysis is more generally applicable at any point of the parameter space, including any points along training. All arguments in Section \ref{Sec:Goldilocks} remain valid because the Hessian exhibits the "bulk+outlier" eigenspectrum throughout optimization (in fact, as \citet{papyan} shows, it becomes even more pronounced with training) inherited from the H- and G-terms, respectively. All properties derived in Section \ref{Sec:Features} that do not explicitly assume random weights (e.g., initial loss) hold more generally beyond initialization because the adopted random model developed by \citet{emergent} does. In their paper, they empirically confirmed that logit gradients cluster around their corresponding means not only at initialization, but also throughout training and at the solution.

In addition, while we comprehensively studied the behaviors of scaled homogeneous models when optimized by gradient descent, the conclusions might change with adaptive algorithms such as AdaGrad or Adam as these optimizers are known to effectively escape flat loss regions \citep{orvieto}. We note that existing studies propose logit normalization for calibration not only after but also during training to improve convergence and generalization \citep{logit-norm, tempcheck}. Given the connection between model confidence and the Goldilocks zone unveiled in our study, future research may investigate how these methods affect positive curvature or, more generally, the eigenspectrum of the loss Hessian. Finally, our findings demonstrate the need for alternative measures to evaluate the quality of network initializations as positive curvature fails to faithfully capture model performance after training.

\section*{Acknowledgements}
The authors thank Mahalakshmi Sabanayagam, Jonathan Niles-Weed, and Kyunghyun Cho for numerous enlightening discussions. AV and JK acknowledge support through NSF Award 1922658. The Flatiron Institute is a division of the Simons Foundation. This work was supported in part through the NYU IT High Performance Computing resources, services, and staff expertise.

\section*{Impact Statement}
This paper presents work whose goal is to advance the field of Machine Learning. There are many potential societal consequences of our work, none which we feel must be specifically highlighted here.

\bibliography{bibli}
\bibliographystyle{icml2024}

\newpage
\appendix
\onecolumn

\section{Hessian Decomposition: Related Works}
\label{App:GaussNewton}
The Gauss-Newton decomposition in \cref{Eq:GN} is a common entry point for many studies on the Hessian of large neural networks. The Hessian exhibits a ``bulk-outlier'' eigenspectrum with the majority of eigenvalues small and clustered around zero and only a handful of large positive outliers \citep{sagun, gurari, ghorbani}. This decomposition is inherited from the individual spectra of $\mathcal{G}$ and $\mathcal{H}$ with the top and the bulk eigenvalues attributed to these two terms, respectively. A few observations offer an intuitive explanation for this claim. 

First, the G-term can be rewritten in matrix form as $\mathcal{G}(p)=J^{\top}[\text{diag}(p)-pp^{\top}]J$ where $J$ is the logit Jacobian. The matrix $\text{diag}(p)-pp^{\top}$ is positive semi-definite, so $\mathcal{G}$ has non-negative eigenvalues. At the same time, $\mathcal{G}$ computed on a single training sample is at most rank $K-1$, suggesting that the full G-term should be low-rank, especially for overparameterized architectures and small training datasets \citep{pennington}. Moreover, \citet{emergent} find empirically that logit gradients on same-class training samples considerably overlap, which implies that eigenspaces of the G-terms of different data samples align. In support of this claim, \citet{papyan3} shows that the top eigenvectors of the Hessian are attributable to high magnitude gradient class means. This agrees with a common observation that the Hessian has $K$ outlier eigenvalues. Recent studies argue that the corresponding eigenvectors encode the decision boundary of the network, and that more outliers emerge for more complex decision boundaries (e.g., after training from an adversarial initialization) \citep{adversarial, alignment}. This observation can potentially correspond with a three-level hierarchical decomposition of the G-term demonstrated by \citet{papyan}, which reveals additional clusters of smaller outliers.

The H-term, on the other hand, has a continuous bulk of small eigenvalues, as independently verified by a number of studies \citep{papyan2}. Assuming that $\mathcal{H}$ is a Wigner random matrix, \citet{pennington} find its limiting spectral density to be consistent with experiments on randomly-initialized ReLU networks without biases. They also note that the H-term has a block off-diagonal structure for ReLU networks, which in particular implies that it has zero excess of positive curvature. In the context of Neural Tangent Kernel (NTK), a kernel-based model for optimization dynamics, logits depend linearly on model parameters, and so $\mathcal{H}$ is identically zero \citep{ntk}.

This pervasive analytical and empirical evidence suggests that, normally, the G-term has a pronounced bulk+outlier eigenspectrum that dominates the bulk eigenvalues found in the H-term. \citet{pennington} note that the eigenspaces of these two matrices are in a nearly generic position and do not align in any special way. To this end, spiked matrix theory suggests that the Hessian eigenspectrum should be well approximated by the combination of individual eigenspectra of $\mathcal{G}$ and $\mathcal{H}$ \citep{spiked}. Since outlier eigenvalues of the Hessian are particularly important, researchers often drop the H-term from \cref{Eq:GN} and focus on the G-term alone \citep{sagun, emergent, eos, li22}.

\section{Eigenvalues of $\mathbb{E}\mathcal{G}$}
\label{App:EigenvaluesEG}

Under the assumptions of the random model on logit gradient clustering described in Section \ref{Sec:Features}, \cref{Eq:EG} offers a clear picture of the eigenstructure of the expected G-term. In particular, it consists of a constant bulk plus $K-1$ outliers:
\begin{align}
\lambda_i(\mathbb{E}_e\mathcal{G}) =
\begin{cases}
(1-\lVert p\rVert^2)\sigma_e^2+d\sigma_c^2\tilde{\lambda}_i & \text{ if } i\leq K-1,\\
(1-\lVert p\rVert^2)\sigma_e^2 & \text{ otherwise. }
\end{cases}\nonumber
\end{align}
where $\tilde{\lambda}_i$ is the $i$-th largest eigenvalue of the matrix $\text{diag}(p)-pp^{\top}$. Since this is a rank-$1$ update to a diagonal matrix, its eigenvalues are interlaced with those of $\text{diag}(p)$; more precisely, for all $i\in[K-1]$, $p_{k_i}\geq\tilde{\lambda}_i\geq p_{k_{i+1}}$ where index $\{k_i\}_{i=1}^{K}$ sorts elements of $p$ in non-increasing order \citep{bunch}. This, for example, immediately proves that when $p$ is uniform, all $K-1$ outlying eigenvalues of $\mathbb{E}_e\mathcal{G}$ are identically $1/K$ as they get squeezed between values $p_k$. The same ``bi-level'' eigenstructure emerges when the expected G-term is taken across multiple training samples, as long as their softmax output remains uniform, as is the case for completely unconfident models ($\alpha\ll 1$). 

\section{The Limiting Behavior of $\mathbf{\text{diag}(p)-pp^{\top}}$}
\label{App:Limit}
Recall that the positive semi-definite matrix $\text{diag}(p)-pp^{\top}$ of rank $K-1$ defined for some probability distribution $p$ over $K$ classes is identically zero when $p$ is one-hot. In this section, we derive a few limiting properties of this matrix as the distribution $p$ gets colder. To this end, for any $\epsilon>0$ and $2\leq S\leq K$, let $p(S, \epsilon)$ be defined as $p_k = \epsilon$ for $k\in[S-1]$, $p_S = 1-(S-1)\epsilon$, and zero otherwise. In other words, $p(S, \epsilon)$ has $S$ non-zero entries with $S-1$ of them equal to $\epsilon$. 

\paragraph{Excess of positive eigenvalues $\mathbf{\Gamma_p}$.} In Section \ref{Sec:Features}, we defined $\Gamma_p$ to be the excess of positive eigenvalues of the matrix $\text{diag}(p)-pp^{\top}$. By inspecting the elements of this matrix, we find
\begin{align}
\label{Eq:Gamma_sq}
\Gamma^2_p = \left[\frac{\text{Tr}(\text{diag}(p)-pp^{\top})}{\lVert\text{diag}(p)-pp^{\top}\rVert_F}\right]^2=\frac{\left(\sum_{k=1}^Kp_k-\sum_{k=1}^Kp_k^2\right)^2}{\sum_{k=1}^K(p_k-p_k^2)^2+\sum_{k\ne c}p_k^2p_c^2}.
\end{align}
Note that $\Gamma_p$ is undefined when $p$ is one-hot, but we can still analyze its limiting behavior. Rewriting \cref{Eq:Gamma_sq} for a distribution $p(S,\epsilon)$ defined above, we get
\begin{align}
\Gamma^2_{p}(S, \epsilon) = \frac{\epsilon^2(4S^2-8S+4)+\mathcal{O}(\epsilon^3)}{\epsilon^2(S^2+S-2)+\mathcal{O}(\epsilon^3)}\quad \Longrightarrow \quad \lim_{\epsilon\rightarrow 0 }\Gamma^2_{p}(S, \epsilon) = 4\frac{S-1}{S+2}.
\end{align}
While this limit evaluates to $3$ when $S=10$ (the number of classes in the datasets considered in this study), this is typically not how softmax output $p$ collapses as we increase logit variance. Assuming no two logits are identically the same, we should eventually get $p=p(2,\epsilon)$ with the two largest logits associated with non-zero softmax probabilities. In this realistic scenario, the limiting value of $\Gamma_p^2$ is $1$, as observed in Figure \ref{Fig:Analysis}. Note that some trajectories of $\Gamma_p^{-2}$ over the entropy of the underlying softmax distribution $p$ seem to converge to a smaller number before eventually reaching $1$. These trajectories correspond to logits with several close high values, so that more than two entries in the softmax output remain positive as we increase the initialization scale.

\paragraph{The eigenspectrum.} As argued above, just before the softmax output $p$ succumbs to the increasing logit variance and collapses to a one-hot vector, it necessarily has the form $p(2,\epsilon)$. In this case, up to a rearrangement of some rows and columns, 
\begin{align}
\text{diag}(p)-pp^{\top} =
\begin{bmatrix}
\epsilon & -\epsilon & \cdots & 0\\
-\epsilon & \epsilon & \cdots & 0\\
\vdots & \vdots & \ddots & \vdots\\
0 & 0 & 0 & 0\\
\end{bmatrix} + \mathcal{O}(\epsilon^2)I_K.
\end{align}
It is easy to see that, neglecting the $\mathcal{O}(\epsilon^2)$ terms, this matrix has a single non-zero eigenvalue equal to $2\epsilon$. In Appendix \ref{App:PreCollapse}, we present some interesting observations about the corresponding eigenvector.

\begin{figure}[H]
\centering
\includegraphics[width=0.6\linewidth]{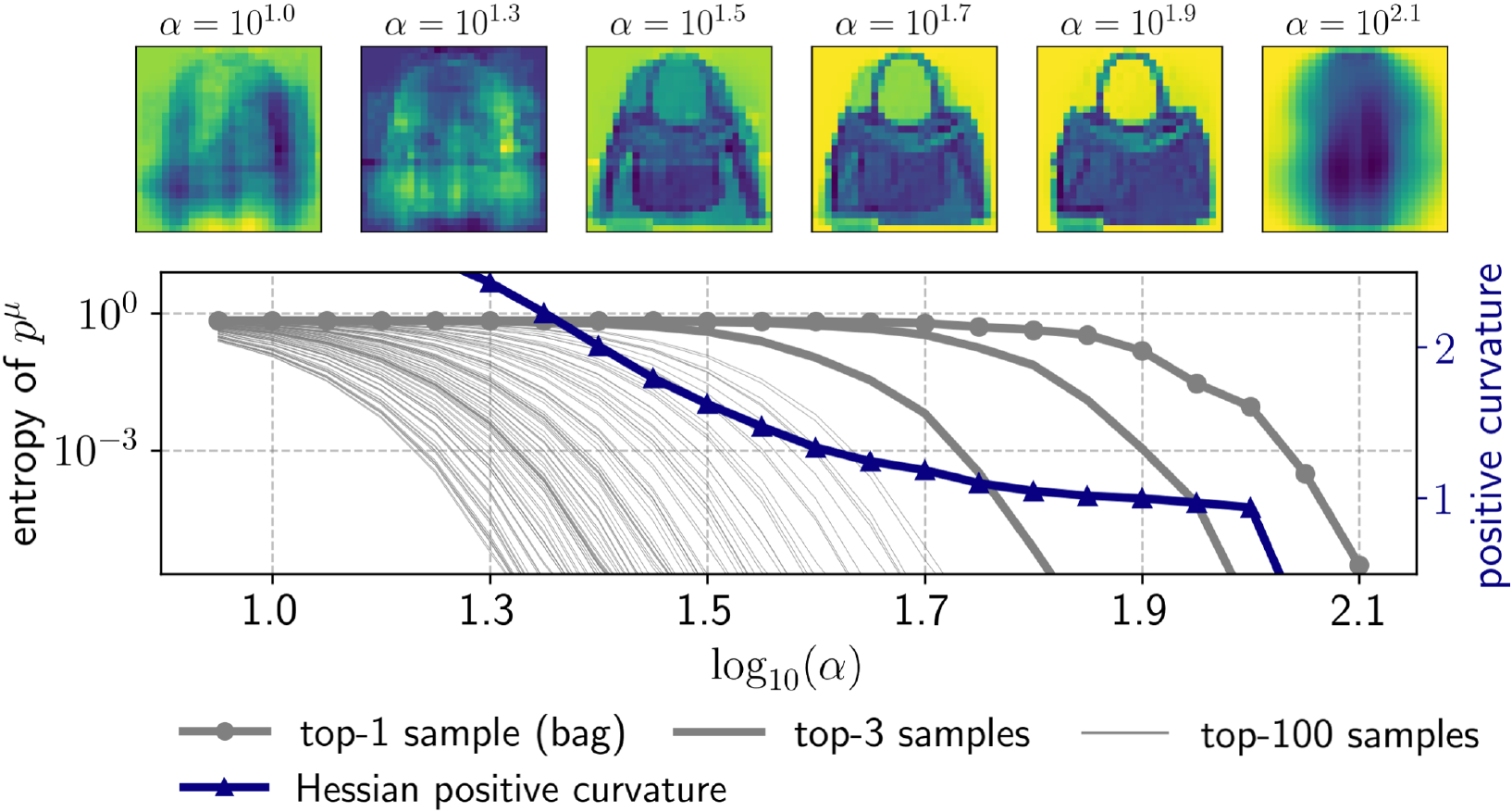}
\caption{The pre-collapse regime of LeNet-300-100 on FashionMNIST. \textbf{Top row:} the top eigenvector of the first layer's Hessian is reshaped into the weight matrix dimensions ($300\times 784$), averaged across the hidden dimension ($300$), and reshaped as an input image ($28\times 28$). Note that the above heatmaps have different scales. As we gently increase $\alpha$, only one input sample (a picture of a bag in this case) remains uncertain, taking full control of the G-term and representing the principle curvature direction for the input layer. \textbf{Bottom:} The collapse of per-sample softmax entropies $\mathbb{H}(p^{\mu})$ and the corresponding positive curvature of the loss.}
\label{Fig:Shoe}
\end{figure}

\section{The Pre-collapse Regime}
\label{App:PreCollapse}

Having examined positive curvature of low-confident models in Section \ref{Sec:Features}, we focus our analysis on the other end of the Goldilocks zone associated with high model confidence. Recall that excess of positive eigenvalues of a positive semi-definite matrix is lower bounded by $1$. Figure \ref{Fig:TrueGoldilocks} illustrates this property for the G-term: as we increase $\alpha$, $\text{Tr}(\mathcal{G})/\lVert \mathcal{G}\rVert_F$ approaches $1$, remains level for a short time (we call this \emph{the pre-collapse regime}), and then suddenly drops to $0$. This property even transfers to the full Hessian, as seen from Figure \ref{Fig:TrueGoldilocks}. Further, in the pre-collapse regime of fully-connected architectures, we mysteriously observe that the principle curvature direction of the first layer's Hessian aligns with one particular sample from the dataset (Figure \ref{Fig:Shoe}). It turns out that we have all the necessary tools to explain this surprising behavior. As we increase $\alpha$, the G-terms $\mathcal{G}^{\mu}$ of individual samples approach zero because $\text{diag}(p^{\mu})-p^{\mu}{p^{\mu}}^{\top}$ does; however, they do so at different rates. When $\alpha$ falls within a very specific narrow range, there is only one sample $\mu_0$ (the least confidently classified one) left with a large enough G-term, so that $\mathcal{G} = \frac{1}{N}\sum_{\mu}^N\mathcal{G}^{\mu} \approx \mathcal{G}^{\mu_0}$. At this point, the top Hessian eigenspace comes from that of $\mathcal{G}^{\mu_0}$ alone, which is spanned by the corresponding logit gradients. Since logit gradients with respect to parameters in the input layer are just the input vector itself, it manifests as the top eigenvector of the Hessian. Furthermore, in Appendix \ref{App:Limit}, we show that $\mathcal{G}(p)$ has only one non-zero eigenvalue when $p$ is on a knife edge from collapsing to a one-hot vector, so that excess of positive eiegnvalues of $\mathcal{G}^{\mu_0}$ and, hence, of the full Hessian, approaches $1$ in the pre-collapse regime.

\section{More Connections to Optimization}
\label{App:Extended}

In this section, we elaborate on the phenomena summarized in Section \ref{Sec:Trainability}. In particular, we present more details about the optimization behaviors observed when scaled homogeneous networks are trained by gradient descent.

\begin{wrapfigure}{r}{0.5\textwidth}
\includegraphics[width=\linewidth]{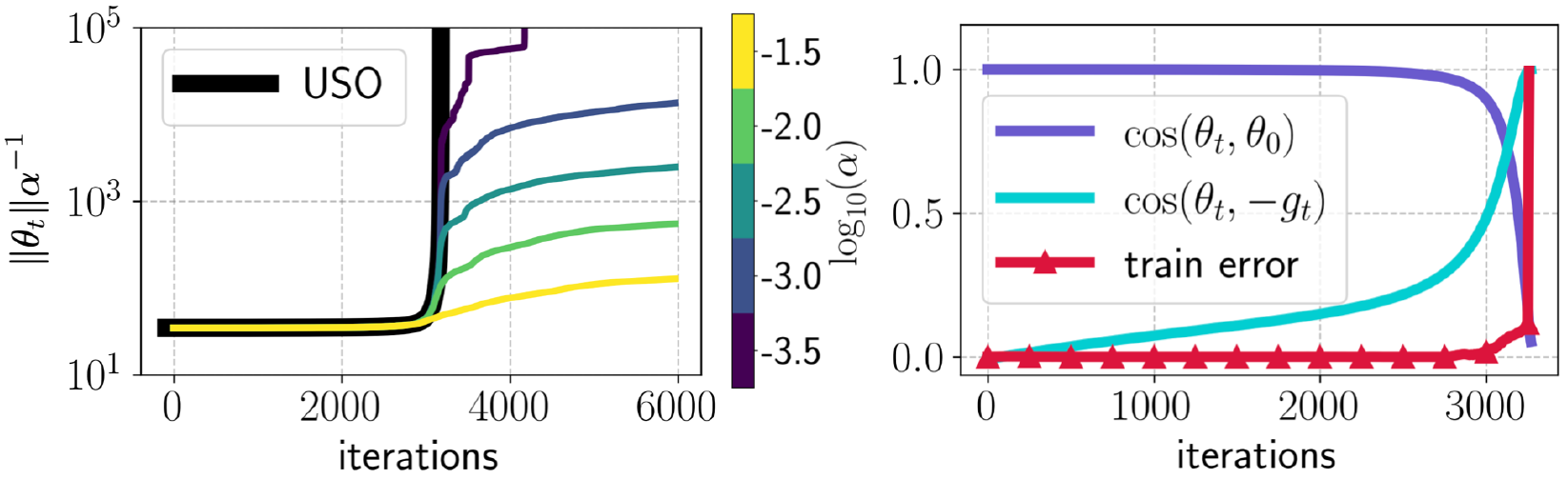}
\caption{Training dynamics of the USO model and networks with low initialization norm ($\alpha<1$), all for LeNet-300-100 on FashionMNIST and $\eta_0=0.0002$. \textbf{Left}: The divergence paths of networks with $\alpha<1$ that initially have a uniform softmax output. The parameter magnitudes of these networks are shown after rescaling back by $\alpha^{-1}$ for a fair comparison with the USO model. The $\alpha$-scaled networks follow the same exact divergence path as the USO model until their weights grow large enough to make softmax output non-uniform. \textbf{Right:} Properties of the divergence (USO model): cosine similarity between the current parameter vector $\theta_t$ and (1) the initial point $\theta_0$, (2) current negative loss gradient $-g_t$. The red curve represents the training error achieved when training LeNet-300-100 initialized at the parameter vector of the current USO model rescaled to an appropriate magnitude: $\lVert\theta_0\rVert\theta_t\lVert\theta_t\rVert^{-1}$ where $\theta_0$ is the original Kaiming initialization.}
\label{Fig:USO}
\end{wrapfigure}

\paragraph{Low-confidence initializations ($\alpha<1$).} Initially, the softmax output of these models is numerically uniform for sufficiently small $\alpha$ due to vanishing logit variance. We observe that, as long as it stays uniform, the model remains in a divergence regime characterized by the increase of parameter norm $\lVert\theta_t\rVert$, gradient norm $\lVert g_t\rVert$ and the alignment $\cos(\theta_t, -g_t)$. Similarly, \citet{omnigrok} report parameter growth during the early training stage of down-scaled networks (see Figure 1 in their paper). Once the parameter norm grows large enough for the prediction to become non-uniform, networks either begin training normally or, if the learning rate is too high, continue diverging. The left plots in Figure \ref{Fig:cmaps} reveal that the maximum admissible base learning rate $\eta_0$ that yields convergence is linearly related to $\alpha$, forming a linear ``staircase''-like separation between trainable and non-trainable setups.

To understand this relationship, recall that classical optimization theory suggests that, for quadratic loss functions, gradient-based methods converge when $\eta<2/\lVert H\rVert_2$ and diverge otherwise \citep{li22}. This result was confirmed in deep learning with more complex loss functions, although slightly larger learning rates are often admissible, too \citep{admiss, eos}. Assuming that the loss curvature $\lVert H\rVert_2$ is approximately constant when different $\alpha$-scaled networks escape the divergence regime and re-enter the Goldilocks zone, this requirement translates to $\eta_0\alpha^{-1}=2/\mathcal{O}(1)$, which corresponds to the linear separation between trainable and divergent models seen in the left plots in Figure \ref{Fig:cmaps}. This extension of the convergence region well into $\alpha<1$ suddenly ends at sufficiently small $\alpha$. In this case, it is the limited number of epochs that prevents networks with smaller initialization reach the Goldilocks zone. Still, we believe that this phase transition should inevitably emerge at some, possibly smaller, value of $\alpha$ due to the adverse effects of remaining in the divergence regime for arbitrarily long.

To study these adverse effects, we propose a ``uniform softmax output'' (USO) model, which is a normally initialized network that always receives updates as if its softmax output is uniform, regardless of its actual predictions. In other words, on every training iteration, we manually curate the update direction by combining logit gradients using coefficients $p_k=1/K-1\{y=k\}$. Equation \ref{Eq:gradients} then implies that the USO model and the down-scaled networks with uniform output have exactly the same evolution for the same base initialization because they receive identical updates. Figure \ref{Fig:USO} (left) confirms that low-confidence networks evolve according to the USO model until their parameters grow large enough for the softmax output to become non-uniform, at which point they re-enter the Goldilocks zone. On the other hand, the USO model never escapes the divergence regime by design, allowing us to study the properties of scaled models with arbitrarily small $\alpha$ as they continue diverging. Figure \ref{Fig:USO} (right) demonstrates that longer divergence is associated with a more severe rotation of the parameter vector, which significantly affects its trainability even after fixing its magnitude, as evident from the training error shown in red. Thus, we expect that for sufficiently small values of $\alpha>0$, scaled networks cannot train normally by the time they reach the Goldilocks zone. Furthermore, longer divergence entails larger gradients, so that a network may just overshoot the Goldilocks zone altogether unless $\eta_0$ is appropriately adjusted. 

\begin{wrapfigure}{r}{0.5\textwidth}
\centering
\includegraphics[width=\linewidth]{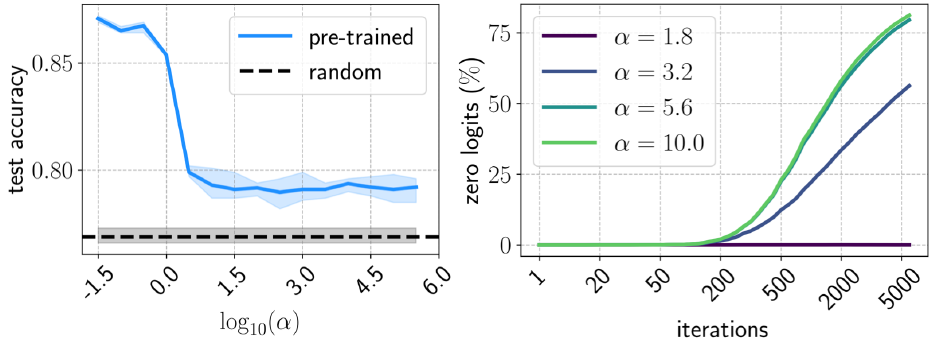}
\caption{\textbf{Left:} Evaluating the features learned by LeNet-300-100 at different initialization scales. We train a linear classifier on data representations extracted from the penultimate layer of scaled networks at convergence (solid) as well as randomly initialized unscaled networks (dashed). Error bands represent min/max across 3 seeds. \textbf{Right:} The proportion of training samples with all logits equal zero as computed by LeNet-5 throughout optimization.}
\label{Fig:Features}
\end{wrapfigure}

\paragraph{High-confidence initializations ($\alpha>1$).} For large-norm initializations, homogeneous networks are overly confident and compute one-hot softmax outputs for all input samples. While both architectures considered in this study exhibit similar optimization patterns for $\alpha<1$, their behavior is remarkably different in this scenario. For LeNet-300-100, we find that predictions remain one-hot throughout training for admissible step sizes where learning is possible. In fact, on every iteration, each training sample requires the learning rate to fall into a very specific narrow range for its softmax output to become non-degenerate, and these ranges are wildly different for different samples. Therefore, it is not possible to tune the step size to escape this extremely confident training regime. Further, note from Figure \ref{Fig:Dynamics} (left) and Figure \ref{Fig:cmaps} (top-left, top-right) that LeNet-300-100 reaches zero training error as long as the learning rate is admissible at initialization. These solutions, however, do not generalize well; the right plot in Figure \ref{Fig:Dynamics} shows that test accuracy of a trained LeNet-300-100 saturates at 75\% across all learning rates shown, which is 10\% lower than that of the unscaled network. This observation suggests that LeNet-300-100 with high-norm initializations adhere to the lazy learning regime in accordance with previous studies even though our optimization methodologies differ \citep{chizat}. To verify this claim, we train a linear classifier on features of the trained $\alpha$-scaled LeNet-300-100 and find them only marginally better than those extracted from a randomly initialized network (see the left plot in Figure \ref{Fig:Features}). At the same time, features learned within the Goldilocks zone exhibit much better generalizability. Thus, even in our setup, gradient descent applied to networks with large initialization can be in the well-studied lazy regime. \citep{overparameterized, du}.

\section{Alternative Estimation of Positive Curvature}
\label{App:Hutchinson}

\begin{wrapfigure}{l}{0.45\textwidth}
\centering
\vspace{-10px}
\includegraphics[width=0.95\linewidth]{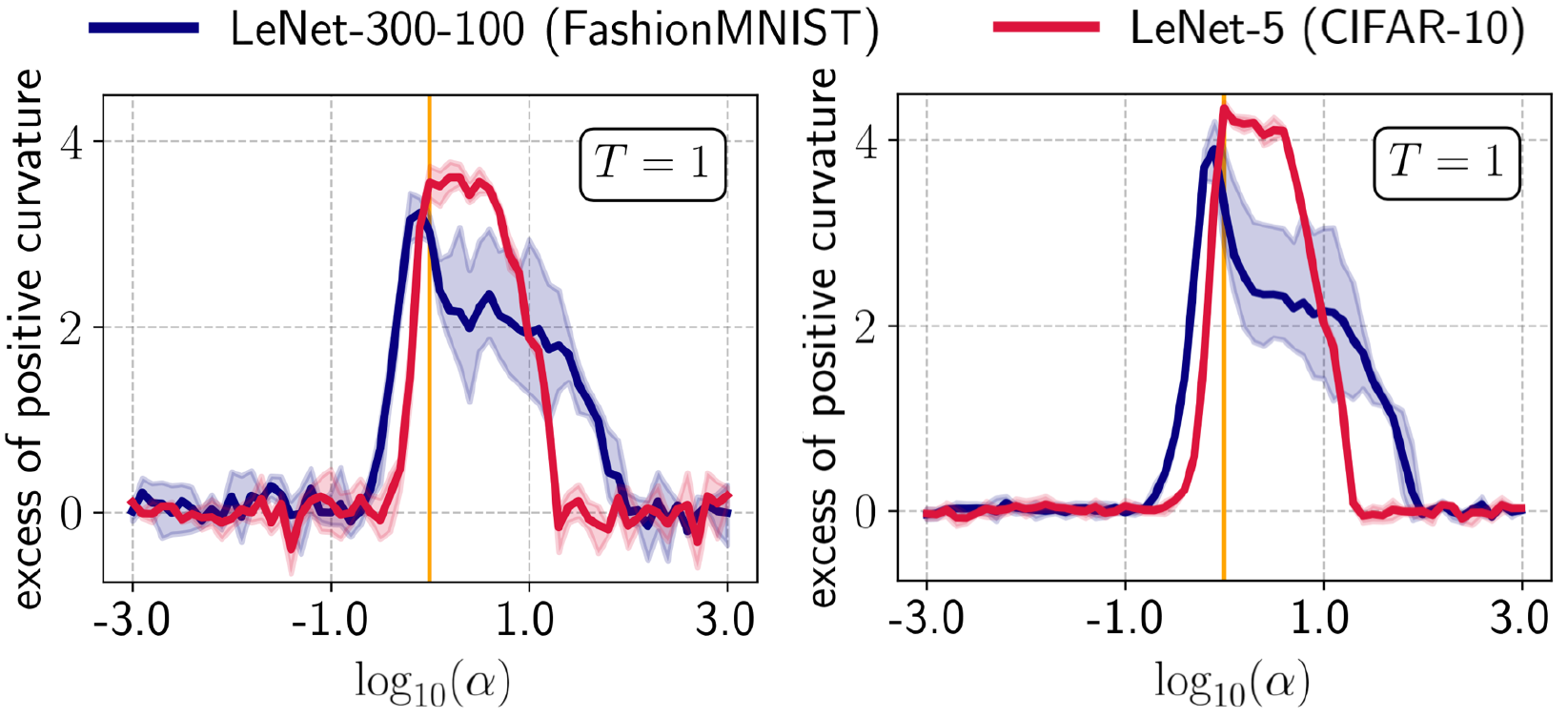}
\caption{Comparing methods for estimating positive curvature of the loss. \textbf{Left:} our method adopted from \citep{goldilocks} \textbf{Right}: Hutchinson's stochastic trace estimation.}
\label{Fig:Hutch}
\end{wrapfigure}

As stated in Section \ref{Sec:Notation}, following the original work of \citet{goldilocks}, we estimate positive curvature of the loss by projecting the Hessian down onto a low-dimensional hyperplane using a sparse orthogonal transformation, and computing the quantity in \cref{Eq:positive-curvature} explicitly for this smaller matrix. Alternatively, we could use the Hutchinson's stochastic trace estimation method to approximate the trace and the norm of the original Hessian $H$ \citep{hutch}. In this approach, we use Hessian-vector products to compute $\text{Tr}(H)\approx \frac{1}{m}\sum_{i=1}^mx_i^{\top}Hx_i$ and $\lVert H\rVert_F^2 \approx \frac{1}{m}\sum_{i=1}^m(Hx_i)^{\top}(Hx_i)$ for $x_i\sim\{+1,-1\}^P$. As Figure \ref{Fig:Hutch} illustrates, Hutchinson's method agrees with the one adopted in this study fairly well, capturing the boundaries and shape of the Goldilocks zone.

\end{document}